\pdfoutput=1

\documentclass[11pt]{article}

\usepackage[final]{acl}

\usepackage{times}
\usepackage{latexsym}

\usepackage[T1]{fontenc}
\usepackage[utf8]{inputenc}
\usepackage{microtype}
\usepackage{inconsolata}

\usepackage{hyperref}       
\usepackage{url}            
\usepackage{booktabs}       
\usepackage{amsfonts}       
\usepackage{nicefrac}       
\usepackage{microtype}      
\usepackage{xcolor}         

\usepackage{algorithm}
\usepackage{enumitem}
\usepackage{amsmath}
\usepackage{mathrsfs}
\usepackage{amssymb}
\usepackage{subfigure}
\usepackage{graphicx}
\usepackage{amsthm}
\usepackage{color}
\usepackage{mathrsfs}
\usepackage{amsfonts}
\usepackage{wrapfig}
\usepackage{hyperref}
\usepackage{blindtext}
\usepackage{caption}
\usepackage{makecell}
\usepackage{algpseudocode}

\usepackage{pifont}
\newcommand{\cmark}{\ding{54}}%

\def\Algnameline{\underline{Na}tural La\underline{n}guage Pr\underline{o}mpt Encapsulation}
\def\Algname{\textit{Capsule} \textit{Prompt}}
\def\Algnameabbr{Nano-Capsulator}
\algnewcommand\algorithmicinput{\textbf{Input:}}
\algnewcommand\INPUT{\item[\algorithmicinput]}
\algnewcommand\algorithmicoutput{\textbf{Output:}}
\algnewcommand\OUTPUT{\item[\algorithmicoutput]}

\title{Learning to Compress Prompt in Natural Language Formats}

\author{
  Yu-Neng Chuang\thanks{~~Work done as an intern at Samsung Research America.} \\
  Rice University \\
  \texttt{ynchuang@rice.edu}\\
  \\ 
  \And
  Tianwei Xing \\
  Samsung Research America \\
  \texttt{t.xing@samsung.com}\\ 
  \And
  Chia-Yuan Chang \\
  Texas A\&M University \\
  \texttt{cychang@tamu.edu}\\ 
  \AND
  Zirui Liu \\
  Rice University \\
  \texttt{zl105@rice.edu}\\ 
  \And
  Xun Chen \\
  Samsung Research America \\
  \texttt{xun.chen@samsung.com}\\ 
  \And
  Xia Hu \\
  Rice University \\
  \texttt{xia.hu@rice.edu}\\ 
  }

\begin{document}
\maketitle
\begin{abstract}

Large language models (LLMs) are excel at processing multiple natural language processing tasks, but their abilities are constrained by inferior performance with long context, slow inference speed, and the high cost of computing the results.
Deploying LLMs with precise and informative context helps users process large-scale datasets more effectively and cost-efficiently.
Existing works rely on compressing long prompt contexts into soft prompts. However, soft prompt compression encounters limitations in transferability across different LLMs, especially API-based LLMs. 
To this end, this work aims to compress lengthy prompts in the form of natural language with LLM transferability. This poses two challenges: (i) Natural Language (NL) prompts are incompatible with back-propagation, and (ii) NL prompts lack flexibility in imposing length constraints.
In this work, we propose a \Algnameline{} (\Algnameabbr{}) framework compressing original prompts into NL formatted \Algname{} while maintaining the prompt utility and transferability. Specifically, to tackle the first challenge, the \Algnameabbr{} is optimized by a reward function that interacts with the proposed semantics preserving loss. To address the second question, \Algnameabbr{} is optimized by a reward function featuring length constraints. 
Experimental results demonstrate that the \Algname{} can reduce \textbf{81.4\%} of the original length, decrease inference latency up to $\mathbf{4.5 \times}$, and save \textbf{80.1\%} of budget overheads while providing transferability across diverse LLMs and different datasets.
\end{abstract}

\section{Introduction}
Large Language Models (LLMs) have demonstrated substantial proficiency across a variety of natural language processing tasks. Despite their significant potential and broad adoption, LLMs are fundamentally limited by the long context length input, which impairs their capability to understand lengthy documents and affects their efficiency during inference~\cite{touvron2023llama, brown2020language, yang2023harnessing, jin2024llm}. As the demand for processing millions of tokens increases, it is progressively crucial to deploy LLMs that are adept at comprehending extended lengths while minimizing budgetary requirements.

\begin{figure}
    \centering
    \includegraphics[width=1.0\columnwidth]{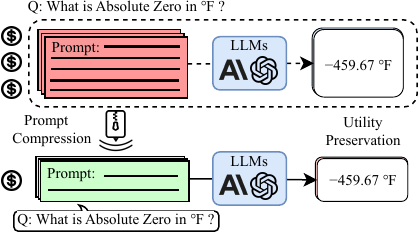}
    \caption{An example of successful prompt compression with NL formats. The compressed NL-formatted prompt (green) aims to obtain a shorter length and maintain transferability and utility of the long prompt (red).}
    \vspace{-0.5cm}
    \label{fig:example}
\end{figure}

To help LLMs better process long context knowledge, recent advancements have focused on compressing long prompt contexts into concise soft prompts. This approach effectively transforms the original extensive prompt into a manageable series of short-length soft prompt tokens. Generally, compression-oriented soft prompts are learned with the guarantee of semantics through self-information~\cite{chevalier2023adapting}, instruction finetuning~\cite{ge2023context, ren2023context}, and the performance alignment via knowledge distillation~\cite{wingate2022prompt, mu2023learning}. However, with the rapid evolution and the growth of API-based accessibility of LLMs, soft prompts pose significant limitations in terms of transferability across different LLMs, implying that well-trained soft prompts can only be effectively adapted to the specific LLMs for which they were designed. This situation creates a critical need to achieve both transferability and utility effectively. A natural question is raised: \textbf{\textit{Can we compress lengthy prompts in a natural language format, yet still preserve utility and ensure transferability among various LLMs?}}

Compressing extended prompts into a shorter, natural language (NL) format continues to be a challenging and unresolved issue. As depicted in Figure~\ref{fig:example}, effective prompt compression entails preserving essential semantic information in a constrained length with successful performance preservation. However, unlike soft prompts, which can be directly optimized with a fixed length, compressing prompts into shorter NL prompts is challenging for several reasons: \textit{(i) NL prompts are incompatible with back-propagation}, as the gradient cannot backward to a discrete raw text; \textit{(ii) NL prompts lack flexibility on imposing strict length constraints}, where overly stringent limitations on generation length may lead to performance degradation. Thus, it is nontrivial to compress lengthy prompts into shorter NL ones.

To tackle the aforementioned problems, we propose a \Algnameline{} (\Algnameabbr{}) framework to effectively compress original prompts into a \Algname{} with the aid of a rewarding technique. Our proposed \Algnameabbr{} aims to encapsulate long prompts into shorter ones under specific generation length constraints, maintaining performance through an explicit semantic preservation objective with reward scores. Specifically, we compress our prompt by employing a semantics-preserving summarization, and then monitor the optimization process using reward scores that reflect the remaining information relevant to the downstream task.
Notably, shorter \Algname{}\textit{s}, characterized by their concise NL formatting, preserve transferability and utility across diverse LLMs. \Algname{} enables two advantages: the preservation of prompt transferability and utility across different LLMs, and the reduction of inference time and budget overheads.  Additionally, \Algnameabbr{} can be directly applied to unseen datasets without any further training, provided these new datasets encompass downstream tasks with similar domains.

To assess the effectiveness of \Algnameabbr{}, we conduct compression experiments with two different prompt types: few-shot demonstration chain-of-thoughts (CoT) and passage prompts of reading comprehension (i.e., content passages). \Algname{} exhibits strong transferability across different LLMs and similar but unseen downstream datasets.
This enables effective adaptation without retraining the prompt compressor.

Our main contributions are concluded as follows: \textit{First}, we introduce and formalize \Algnameabbr{} framework, which can effectively generate high-quality \Algname{} with prominent transferability across multiple LLMs and unseen datasets with similar downstream tasks.
\textit{Second}, we effectively reduce the original prompts to \textbf{81.4\%} of their initial length and transform them into NL-formatted \Algname{}, which retains the prompt's transferability and utility across various LLMs. Our compression mechanism significantly decreases up to $\mathbf{4.5 \times}$ of the inference latency and saves \textbf{80.1\%} of the budget overheads for input sequences.
\textit{Third}, experimental results demonstrate that \Algname{} can efficiently perform across diverse LLMs, which is applicable to both few-shot demonstration CoT and input contextual prompts.





\section{Related Work} 
\subsection{Soft Prompt Compression} In the realm of prompt compression for LLMs, most of 
the existing work aims to compress the prompt into soft prompts. Soft prompts are trainable vectors that are optimized in conjunction with a designated LLM, which embeds the original content information of the long hard prompts into low-dimensional vectors. 

The first line of work~\cite{wingate2022prompt} leverages the knowledge distillation object to extract the information from hard prompts to soft prompts. The compressed soft prompts are expected to capture high-level concepts and preserve the fluency from the original hard prompts.
The second line of work~\cite{chevalier2023adapting} employs the summarization capabilities of LLMs to condense lengthy and complex prompts into soft prompts. The process involves dividing the input prompts into multiple segments and sequentially compressing the information from the original prompt into smaller segments of soft prompts, where they assemble from these individual fragments of soft prompts to form the final soft prompts.
Another work, Gist Token~\cite{mu2023learning}, condenses instruction prompts into customized prefix soft prompts by training a virtual soft prompt predictor. 

Nevertheless, the transferability of soft prompt-based compression across diverse LLMs is constrained, necessitating the re-training of soft prompts with each change in the specified LLMs. This means that the soft prompts generated are specifically tailored to work only with that particular LLM, which falls short in maintaining transferability across different LLMs, especially applied on API-based LLMs, such as Claude2~\cite{claude} and PaLM~\cite{vertex}.

\subsection{Context Distillation for Compression} 
Besides directly compressing hard prompts into soft prompt vectors, recent advancement~\cite{li2023compressing, jiang2023llmlingua} involves computing the self-information scores or perplexity of the given input context prompt to shorten the prompt length. This process includes filtering out words with lower scores from the input prompt, resulting in a more concise input during the inference stage. The primary distinction between our work and these recent studies is that they operate prompt compression without considering any information from downstream tasks, resulting in inferior performance while directly applying to downstream tasks or transferring between similar but unseen downstream datasets.

\section{Long Prompt Encapsulation} 
\begin{figure*}
    \centering
    \includegraphics[width=2\columnwidth]{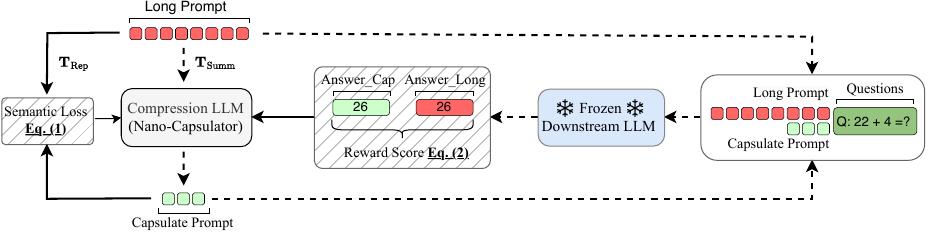}
    \caption{The illustration of \Algnameabbr{} training framework. \Algnameabbr{} compress the long prompt with the action of semantic (Equation~\ref{eq:sem}) and utility preservation (Equation~\ref{eq:reward}). Questions are sampled from the training set to develop the reward scores for utility preservation.}
    \label{fig:over}
\end{figure*}

We systematically introduce the \Algnameabbr{} framework in this section. Figure~\ref{fig:over} illustrates the overall pipeline of \Algnameabbr{}. In particular, our pipeline initially compresses text into  NL-formatted \Algname{} and concurrently optimizes their utility using the proposed rewarding method. The design of the NL-formatted compression aims to maintain prompt utility and preserve transferability among different LLMs.

\subsection{Prompt Encapsulation}
The primary aim of \Algnameabbr{} is to preserve the inherent utility of the original pre-compressed text and ensure the compressed prompt closely reaches the designated length constraint. Specifically, the prompt encapsulation mechanism includes two components to effectively generate \Algname{}: (1) NL-formatted prompt compression and (2) prompt utility preservation. The learning of \Algnameabbr{} involves integrating two components and optimizing them concurrently, thereby assuring the compressed prompts are sufficient to preserve their inherent utility.

\subsubsection{NL-formatted Prompt Compression} We adopt an unsupervised training approach featuring semantic preservation loss, motivating the model to compress contexts while retaining similar semantic content. In this work, we shorten the long prompts by summarizing their context and applying our proposed semantics loss $\mathcal{L}_{\text{Comp}}$ to ensure maximal preservation of semantic meaning.
Here, semantics refers to the logical thinking process from the few-shot demonstration CoT and the beneficial content from context passages.

Given the original prompt $K = \{ k_1, \cdots k_n\}$ consisted of $n$ tokens to the \Algname{} $C = \{ c_1, \cdots c_m\}$ with $m$ tokens, where $n \gg m$. 
Our semantics loss aims to ensure the maximal semantics preservation by measuring the similarity between the hidden state embedding of $C$ and $K$ of \Algnameabbr{} $\mathcal{F}(\cdot~|~\theta_{C})$. To obtain the hidden state embedding of $K$, we instruct $\mathcal{F}(\cdot~|~\theta_{C})$ to replicate the input prompt $K$, which aids in better preserving and embedding the semantic meanings of $K$.
Specifically, $d$-dimensional hidden state embedding of $K$ and $C$ can be generated by $\boldsymbol{e}_K \sim P_{\mathcal{F}}(K~|~ \theta_{C}, \textbf{T}_{\text{Rep}})$ and $\boldsymbol{e}_C \sim P_{\mathcal{F}}(C ~|~ \theta_{C}; \textbf{T}_{\text{Summ}})$, where $\textbf{T}_{\text{Rep}}$ and $\textbf{T}_{\text{Summ}}$ denote a replicating instruction and a summarizing instruction, respectively.
With the aid of $\textbf{T}_{\text{Rep}}$, we compel $\mathcal{F}(\cdot~|~\theta_{C})$ to replicate $K$ under the model parameter $\theta_{C}$, ensuring that $\boldsymbol{e}_K \in \mathbb{R}^d$ accurately represents the embedding of $K$. Following the criterion, $\mathcal{F}(\cdot~|~\theta_{C})$ essentially minimizes the semantics loss as follows:
\begin{equation}
    \mathcal{L}_{\text{Comp}} = \mathbb{E}_{C}
    \big[
    D_\text{dist}(\boldsymbol{e}_K ~||~ \boldsymbol{e}_C)
    \big]
    \label{eq:sem}
\end{equation}
where $D_\text{dist}(\cdot ~||~ \cdot)$ can be any suitable distance measurement in metric space. In this work, we leverage mean square error as the distance function to measure the similarity between $\boldsymbol{e}_K$ and $\boldsymbol{e}_C$.

\begin{algorithm}[t!]\small
\caption{Algorithm of \Algnameabbr{}}
\label{alg:nano}
\begin{algorithmic}[1]
\INPUT Original Long Prompt $K$, Compression Instructions $\textbf{T}_{\text{Rep}}$ and $\textbf{T}_{\text{Summ}}$, and pre-trained frozen LLMs $\mathcal{G}^{*}(\cdot)$, Sampled set of downstream task questions $Q$.
\OUTPUT NL-formatted Capsule Prompts $C$.

\State Initialize $\mathcal{F}(\cdot~|~\theta_{C})$ and $\mathcal{G}^{*}(\cdot)$ with pre-trained weights 
\While{not convergence}
\State Generate $C$ from $\mathcal{F}(K~|~\textbf{T}_{\text{Rep}}, \textbf{T}_{\text{Summ}}, \theta_{C})$
\State Randomly sample a set of questions $Q$
\State Receive reward scores from $\mathcal{R}_{\text{cap}}(\mathcal{G}^{*}(\cdot), Q, C, K)$
\State $\mathcal{F}(\cdot~|~\theta_{C}) \gets$ minimizing with $\mathcal{L}_{\text{Nano}}(\cdot)$
\EndWhile
\end{algorithmic}
\end{algorithm}

\subsubsection{Prompt Utility Preservation} To impose a constraint on the generated length while preserving utility, we establish a reward function $\mathcal{R}_{\text{cap}}(\cdot)$ featuring a strict cut-off mechanism $\Phi(\cdot)$ for restricting the generated length of \Algname{}. The high-level idea of the reward function is to calculate the score changes of the downstream task question based on leveraging the original prompt $K$ and the \Algname{} $C$. Notably, the reward function employs a truncation strategy to limit the $C$ to a predetermined length before proceeding to compute the scores using the reward function. In this manner, the \Algname{} that surpasses the specified length threshold could be assigned a lower reward score as a result of the cut-off mechanism.

Formally, given $K$ and $C$ along with arbitrary pre-trained frozen LLMs $\mathcal{G}^{*}(\cdot)$ and a sampled set of downstream task questions $Q$, the reward function $\mathcal{R}_{\text{cap}}(\cdot)$ can be defined as follows:
\begin{equation}
    \mathcal{R}_{\text{cap}} = \mathbb{E}_{Q}
    \big[~
    \mathbf{I}\{~
        \mathcal{G}\big(\Phi(C_i) \oplus Q_i\big) ~||~ \mathcal{G}\big(K_i \oplus Q_i\big)
    ~\}
    ~\big]
    \label{eq:reward}
\end{equation}
where $\mathbf{I}(\cdot ~||~ \cdot)$ denotes an arbitrary reward metric for yielding the reward score., and $\oplus$ represents concatenation of prompts and questions. In this study, we calculate the reward scores using the mean square error between the hidden state embedding from $\mathcal{G}^{*}(\cdot)$. It's noteworthy that $\mathbf{I}(\cdot ~||~ \cdot)$ can be replaced by other metrics, such as accuracy and GPT4Eval Scores~\cite{liu2023g}, facilitating its potential application to API-based LLMs.

\subsubsection{Compression with Reward}
Upon receiving the reward scores from $\mathcal{R}_{\text{cap}}$ as per Equation~\ref{eq:reward}, we synchronize these scores with the semantic loss $\mathcal{L}_{\text{Comp}}$ to maintain utility. 
Formally, the ultimate objective function of \Algnameabbr{} can be indicated as:
\begin{align}
    \mathcal{L}_{\text{Nano}} = \mathcal{L}_{\text{Comp}}(\cdot|\theta_C) * \mathcal{R}_{\text{cap}}(\cdot|\theta^*)
    \label{eq:nanoobj}
\end{align}
where $\theta^*$ denotes the frozen model parameters of $\mathcal{G}^{*}(\cdot)$ and $\theta_C$ is the trainable parameters of \Algnameabbr{}.
The fundamental principle of $\mathcal{L}_{\text{Nano}}(\cdot)$ is to impose penalties when shorter versions of \Algname{} exhibit inferior performance. This implies that if a \Algname{} receives a low reward score from Equation~\ref{eq:reward}, its semantic loss will be composed by a high penalty value, resulting in a substantial semantic loss value as punishment during the training phase of \Algnameabbr{}.

\subsection{Algorithm of \Algnameabbr{}}
The framework of \Algnameabbr{} is detailed in Algorithm~\ref{alg:nano}. \Algnameabbr{} adheres to Equation~\ref{eq:sem} for the preservation of semantic meaning and integrates the rewarding function from Equation~\ref{eq:reward} to maintain the utility of compressed NL-formatted prompts. The two elements are then aligned, as depicted in Equation~\ref{eq:nanoobj}, and optimized simultaneously with the goal of obtaining compressed NL-formatted prompts of high utility. In the inference phase, \Algnameabbr{} is solely required to produce the compressed version of \Algname{} from the provided long input prompt.

\section{Experiments} 
In this section, we conduct experiments to evaluate the performance of \Algnameabbr{}, aiming to answer the following three research questions:
\begin{itemize}[leftmargin=*]
    \item \textbf{RQ1:} How does \Algnameabbr{} perform in terms of the efficacy and transferability among different LLMs and datasets?
    \item \textbf{RQ2:} How do the two components of \Algnameabbr{} contribute to the compression performance for utility preservation?
    \item \textbf{RQ3:} What are the inference latency and impact factors of \Algname{}?
\end{itemize}

\subsection{Dataset}
We conduct compression experiments with two different prompt types: few-shot CoT and passage prompts of reading comprehension. The details of the datasets are provided as follows:

\vspace{2mm}
\noindent\textbf{Few-shot CoT Dataset.} We choose two reasoning datasets to evaluate the proposed framework.
\begin{itemize}[leftmargin=*, itemsep=0pt, topsep=-1mm]
    \item \textbf{CommonsenseQA~\cite{talmor-etal-2019-commonsenseqa}:} The CommonsenseQA (CSQA) dataset is a publicly accessible collection of multiple choice questions with 1221 samples for the commonsense reasoning task. CSQA presents questions characterized by intricate semantics, typically demanding reasoning grounded in pre-existing knowledge. 

    \item \textbf{GSM8K~\cite{cobbe2021gsm8k}:} The GSM8K is a dataset containing 1319 samples of graduate school math questions. Each question is collected from the Math World Problem Repository~\cite{roy2015solving} with a numerical answer.
\end{itemize}

\vspace{2mm}
\noindent\textbf{Reading Comprehension Dataset.}
\begin{itemize}[leftmargin=*, itemsep=0pt, topsep=-1mm]
    \item \textbf{MultiRC~\cite{MultiRC2018, eraser2019}:} MultiRC (Multi-Sentence Reading Comprehension) comprises a collection of brief paragraphs paired with multi-sentence questions, where the answers can be derived from the paragraph's content. The dataset obtains 24029 samples for training, 3214 samples for validating, and 4848 samples for testing.  

    \item \textbf{TriviaQA LongBench~\cite{2017arXivtriviaqa}:} TriviaQA LongBench (TriviaQA-Long) is a reading comprehension dataset featuring 300 question-answer-evidence triples collected from the Longbench dataset. It includes question-answer pairs created by trivia enthusiasts, along with independently sourced evidence documents, offering robust supervision for responding to the questions.
\end{itemize}

\begin{table*}[t]
\centering
\fontsize{7.5pt}{6pt}\selectfont
\begin{tabular}{l | ccc | ccc | cc | cc}
    \toprule
    & \multicolumn{3}{c}{CSQA} |& \multicolumn{3}{c}{GSM8K} |& \multicolumn{2}{c}{MultiRC} |& \multicolumn{2}{c}{TriviaQA-Long} \\
    \cmidrule(lr){2-4}\cmidrule(lr){5-7}\cmidrule(lr){8-9}\cmidrule(lr){10-11} 
    & Manual & Zero-shot & Ours & Manual & Zero-shot & Ours & Original & Ours & Original & Ours \\
    \midrule
    Vicuna-13B & 60.4 & 44.6 & 58.8 & 34.4 & 25.3 & 31.9 & 57.3 & 57.1 & 86.0 & 88.8\\
    PaLM & 73.7 & 67.5 & 75.5 & 62.8 & 56.8 & 59.5 & 72.7 & 72.2 & 78.9 & 78.8\\
    Claude2 & 76.6  & 69.4 & 74.6 & 85.6 & 52.7 & 84.9 & 59.4 & 58.2 & 95.0 & 92.3 \\
    \midrule\midrule
    Length (\# of Token) & 831 & -- & 154 & 751 & -- & 231 & 378.39 & 95.66 & 915.7 & 422.6 \\
    Compress Rate (\%) & -- & -- & 81.4\% & -- & -- & 69.3\% & -- & 74.71\% & -- & 53.84\% \\
    \bottomrule
\end{tabular}%
\caption{Evaluation of \Algnameabbr{} among different LLMs. The results show that \Algnameabbr{} compress up to 81.4\% of the original long prompt and save up to 80.1\% of the expense on requesting for LLM API calls.}
\vspace{-0.3cm}
\label{tab:mainout}
\end{table*}    

\subsection{Experiment Settings}
In this part, we introduce the experimental settings and metrics for evaluating \Algnameabbr{}. Two distinct types of transferability evaluations are taken into account. \textit{To verify the model transferability}, the compression models are trained on one downstream LLM, and tested on different downstream LLMs. The evaluation is performed on the same dataset, with a division of 70\% allocated for training and validation and 30\% designated as the testing set.
\textit{To assess data transferability}, we train the compression models using one seen dataset and then test them on unseen datasets that the models have not previously encountered with the same downstream tasks.
The considered compression settings and implementation details are shown as follows. Two types of prompt compression tasks are focused on.

\vspace{0.2cm}
\noindent\textbf{Few-shot CoT Compression Task.} For the few-shot CoT compression task, we randomly compile seven examples from the CSQA dataset and eight from the GSM8K dataset following ~\cite{wei2022chain}, all selected from their respective training sets, to construct the few-shot CoT. During the training phase, a total of 1,000 CoT samples are then gathered to serve as the training data for \Algnameabbr{}. During the inference stage, we aim to compress the manual few-shot CoT proposed in~\cite{wei2022chain}, where the demonstrations in manual CoT are eliminated from any training set. The primary evaluation metric used in both CSQA and GSM8K datasets is accuracy, implying that the model scores only when it provides answers that exactly match the expected responses.

\vspace{0.2cm}
\noindent\textbf{Reading Comprehension Compression Task.} For the reading compression task, we aim to compress the reading paragraphs from the question-answer triplets. Due to the limitation of GPU memory, we eliminate the paragraphs that exceed 2k tokens in TriviaQA from the LongBench dataset, resulting in an average length of 900 tokens, while MultiRC remains to utilize the all paragraphs in the dataset. Throughout the training phase, we select 2,000 question-answer-paragraph triplets in the MultiRC dataset to serve as training data and leverage all training data in the TriviaQA-Long dataset. Our framework is evaluated on the entire set of testing data, using accuracy as the metric of assessment.

\vspace{0.2cm}
\noindent\textbf{Implementation Details.} In primary experiments, we utilize Vicuna-7B~\cite{vicuna2023} as the initial compression model $\mathcal{F}(\cdot~|~\theta_{C})$ in \Algnameabbr{}. The pre-trained LLMs $\mathcal{G}^{*}(\cdot)$ is given as Vicuna-7B with frozen weights. We train \Algnameabbr{} using Vicuna-7B and then assess the generated \Algname{} with various LLMs other than Vicuna-7B, in order to evaluate its transferability.
To reduce memory consumption during training, we utilize LoRA\footnote{PEFT: \url{https://github.com/huggingface/peft}} and train the \Algnameabbr{} using two NVIDIA A40 GPUs of 48GB memory. We employ the Adam optimizer for the fine-tuning process, with a learning rate set at 5e-6 under the gradient clipping of 0.8, depending on the datasets. The instructions that leverage for prompt encapsulation $\textbf{T}_{\text{Rep}}$ and $\textbf{T}_{\text{Summ}}$ are listed in Table~\ref{tab:instruct} from Appendix~\ref{apx:prompt}.

\begin{table}[t]
\small
\centering
	\begin{tabular}{l ccc}
        \toprule
        & \multicolumn{3}{c}{Claude2~\cite{claude}} \\
        \cmidrule(lr){2-4}
        Cost(\$) & Original & \Algname{} & Saved \\
        \midrule 
        CSQA & 15.03 & 3.30 & -77.9\% \\
        GSM8K & 5.22 & 1.88 & -63.9\% \\
        MultiRC & 45.91 & 13.01 & -71.6\% \\
        TrivaQA-Long & 2.14 & 0.42 & -80.1\% \\
        \bottomrule
    \end{tabular}%
    \caption{API cost comparison of \Algname{} and original prompt, where \Algname{} save up to 80.1\% of the original cost.}
    \vspace{-0.3cm}
\label{tab:clade_cost}
\end{table}

\subsection{Main Results (RQ1)}
\vspace{0.2cm}
\noindent\textbf{Model Transferability.}
To assess the effectiveness and transferability, we compress the original input prompts into the \Algname{} by \Algnameabbr{}. 
We then evaluate the transferability and utility of these compressed prompts across three different LLMs not included in the pre-training of \Algnameabbr{}: Vicuna-13B~\cite{vicuna2023}, PaLM~\cite{vertex}, and Claude2~\cite{claude}. 
The main findings are presented in Table~\ref{tab:mainout}. In the table, "Manual" refers to the manually created few-shot CoT proposed by~\cite{wei2022chain}, "Zero-shot" denotes the zero-shot CoT followed~\cite{kojima2022large}, and "Original" indicates the original paragraphs used in the reading comprehension tasks. 

In the primary experiment, we establish a compression constraint limiting to a maximum of 150 tokens for the CSQA and MultiRC datasets; and a maximum of 350 and 500 tokens for the GSM8K and TriviaQA-Long dataset, where 150 tokens are not sufficient for preserving the logic of GSM8K and TriviaQA-Long dataset.
We observe that the \Algnameabbr{} obtains up to 81.4\% of the compression rate and saves up to 80.1\% of the Claude2 API cost compared to the original input prompts, as displayed in Table~\ref{tab:clade_cost}. The cost of PaLM API can be found in Appendix~\ref{apx:cost}.
For utility preservation, \Algnameabbr{} retains the original performance on the CSQA, GSM8k, and TriviaQA-Long datasets mostly among three LLMs. Remarkably, \Algnameabbr{} maintains almost identical performance to that achieved with non-compressed prompts in MultiRC datasets. The significant compression rate can advantageously impact the LLMs by allowing for a higher tolerance in batch inference, accompanied by reduced latency and cost.

\begin{figure}[t]
    \centering
    \includegraphics[width=.4\textwidth]{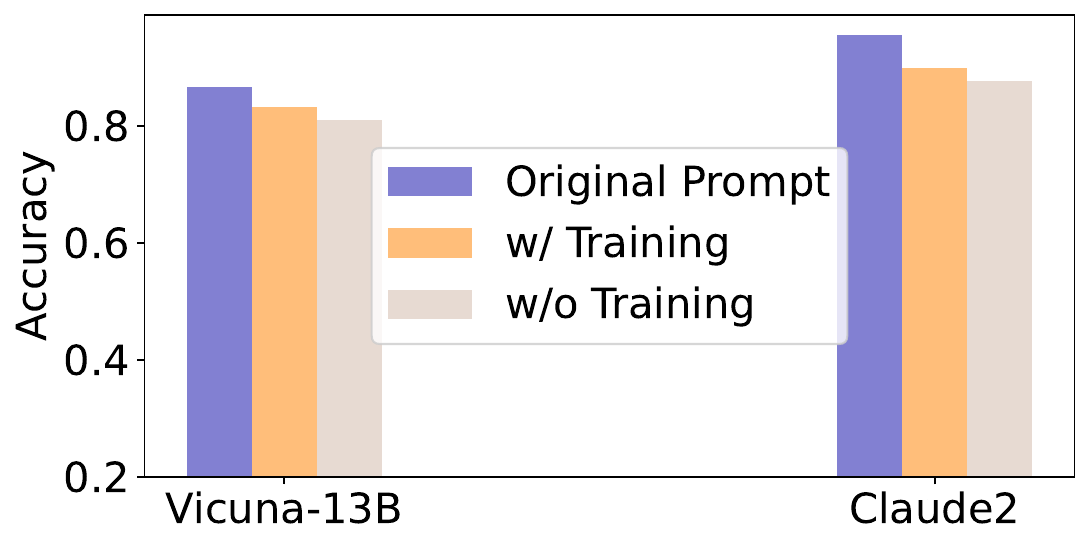}
    \vspace{-0.3cm}
    \caption{Evaluation of transferability on \Algnameabbr{} across unseen datasets.}
    \label{fig:transdata}
    \vspace{-0.3cm}
\end{figure}

\vspace{0.2cm}
\noindent\textbf{Dataset Transferability.}
We previously assessed the effectiveness of \Algnameabbr{} within the same datasets, where the testing set was derived from the same training domain. In this section, we investigate the transferability of \Algnameabbr{} across unseen datasets (i.e., not in training data) with the same downstream tasks.
We train \Algnameabbr{} on the MultiRC dataset (seen dataset) and test on BoolQ~\cite{clark2019boolq} (unseen dataset) without any further training, where BoolQ is also a reading comprehension dataset, under Vicuna-13B and Claude2. The results are demonstrated in Figure~\ref{fig:transdata}. We see a competitive performance with only a slight accuracy drop compared to the training version of \Algname{}. While \Algname{} yields better performance with training, the results indicate that \Algnameabbr{} possesses a great property of data transferability.


\subsection{Contributions of Utility Preservation (RQ2)}

\begin{figure}[t]
\minipage{0.5\columnwidth}
  \includegraphics[width=0.93\columnwidth]{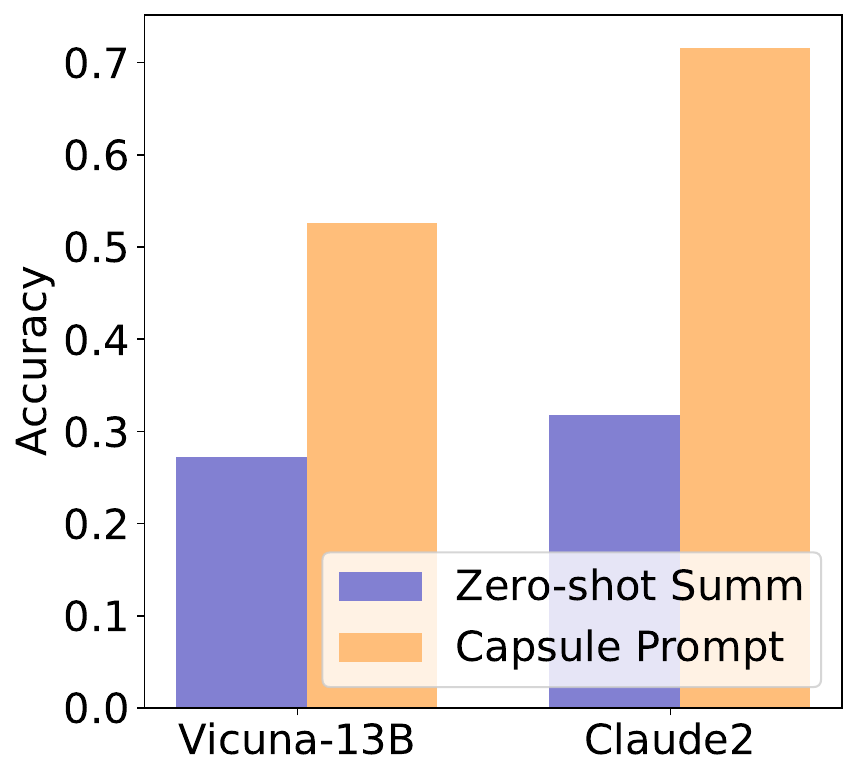}
\endminipage\hfill
\minipage{0.5\columnwidth}
  \includegraphics[width=1.0\columnwidth]{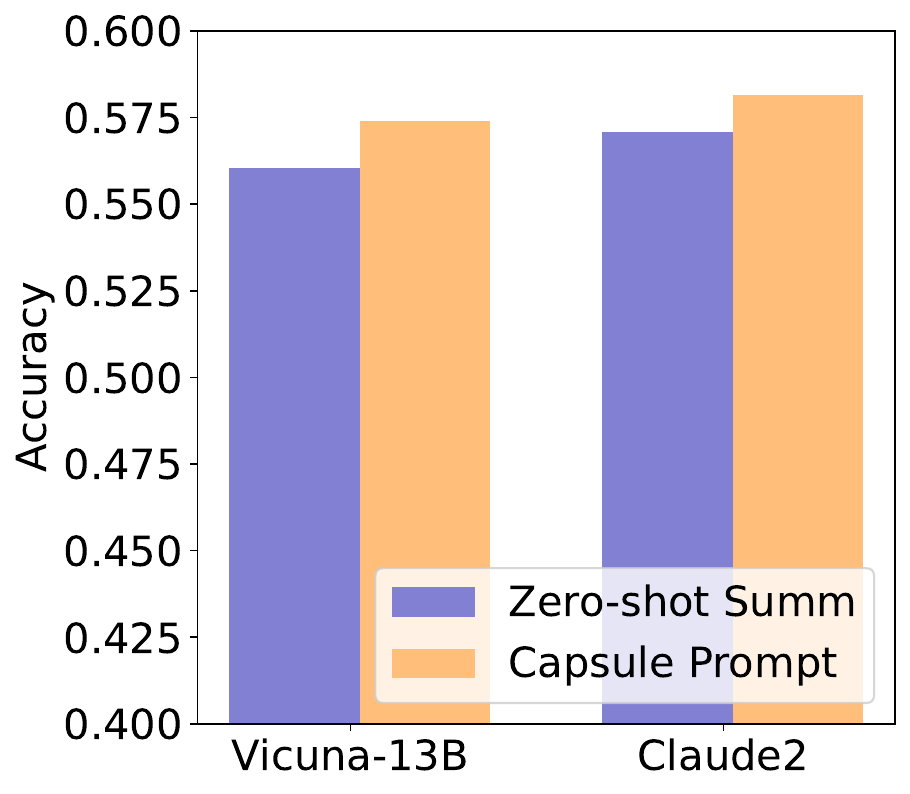}
\endminipage
    \caption{Comparison results of \Algname{} and Zero-shot Summarization on GSM8K dataset (left) and MultiRC dataset (right).}
\vspace{-0.2cm}
\label{fig:zerosumm}
\end{figure}

\begin{figure}[t]
\minipage{0.5\columnwidth}
  \includegraphics[width=1.\columnwidth]{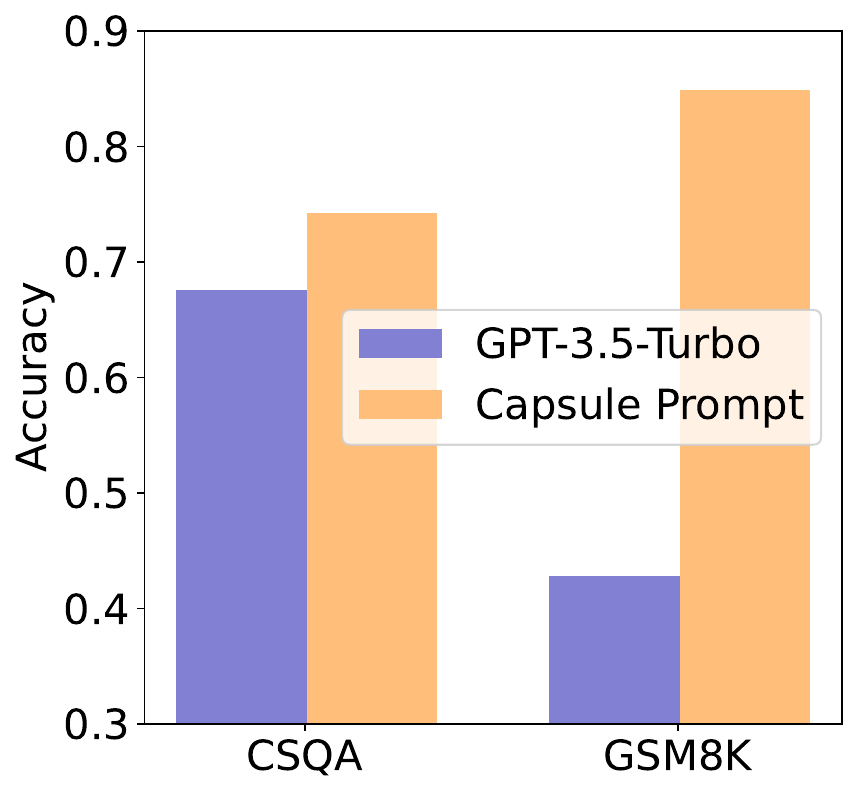}
\endminipage\hfill
\minipage{0.5\columnwidth}
  \includegraphics[width=1.\columnwidth]{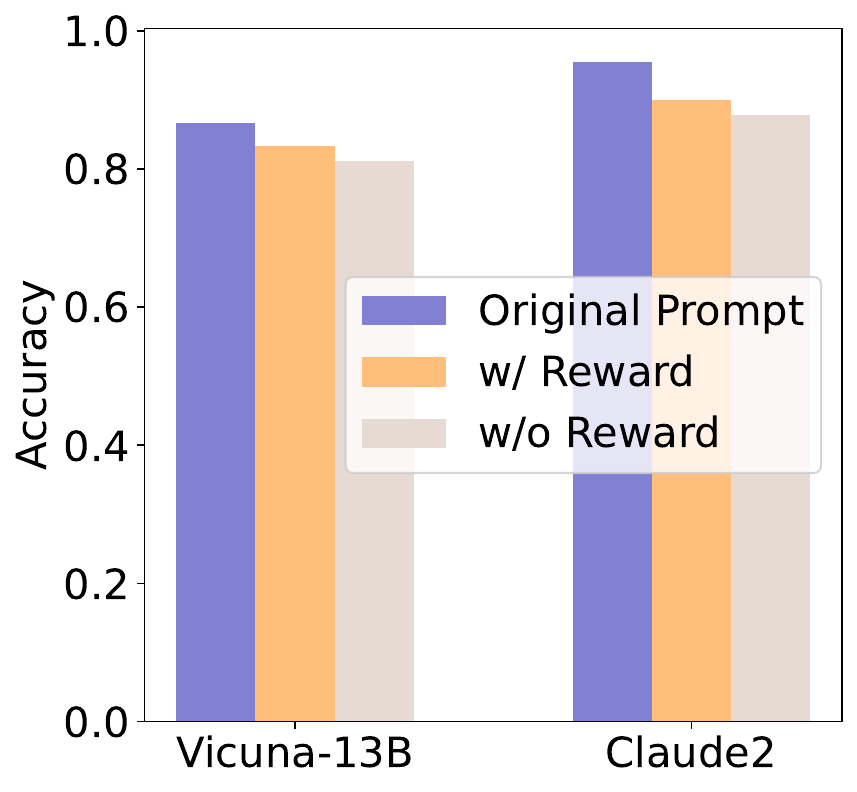}
\endminipage
\vspace{-0.2cm}
\caption{Ablation studies of comparison with \Algname{} and GPT-35-Turbo Summarization on CSQA dataset and GSM8K dataset (left); and of the contribution of Reward Function from Equation~\ref{eq:reward} (right).}
\vspace{-0.3cm}
\label{fig:gptsumm}
\end{figure}

In this section, we explore the effectiveness of components from \Algnameabbr{}. Specifically, we conduct ablation studies from two perspectives. \textit{First}, we evaluate the efficacy of semantic preservation. We compare \Algnameabbr{} with in-context zero-shot summarization generated by Vicuna-7B, as Vicuna-7B is the initial model weight of \Algnameabbr{} for prompt compression. The results are demonstrated in Figure~\ref{fig:zerosumm} with the comprehensive comparison of three LLMs, including Vicuna-13B and Claude2. We observe that \Algname{} yielded by \Algnameabbr{} outperforms in all scenarios, which means that our \Algnameabbr{} can significantly preserve more semantic information and preserve prompt utility. Additionally, we assess the performance by directly employing GPT-3.5-Turbo to summarize the provided prompts. Figure~\ref{fig:gptsumm} (left) illustrates that \Algnameabbr{} maintains a higher level of prompt utility compared to GPT-3.5-Turbo, resulting in enhanced performance on both the CSQA and GSM8K tasks.

\begin{figure}[t]
\minipage{0.5\columnwidth}
  \includegraphics[width=1.0\columnwidth]{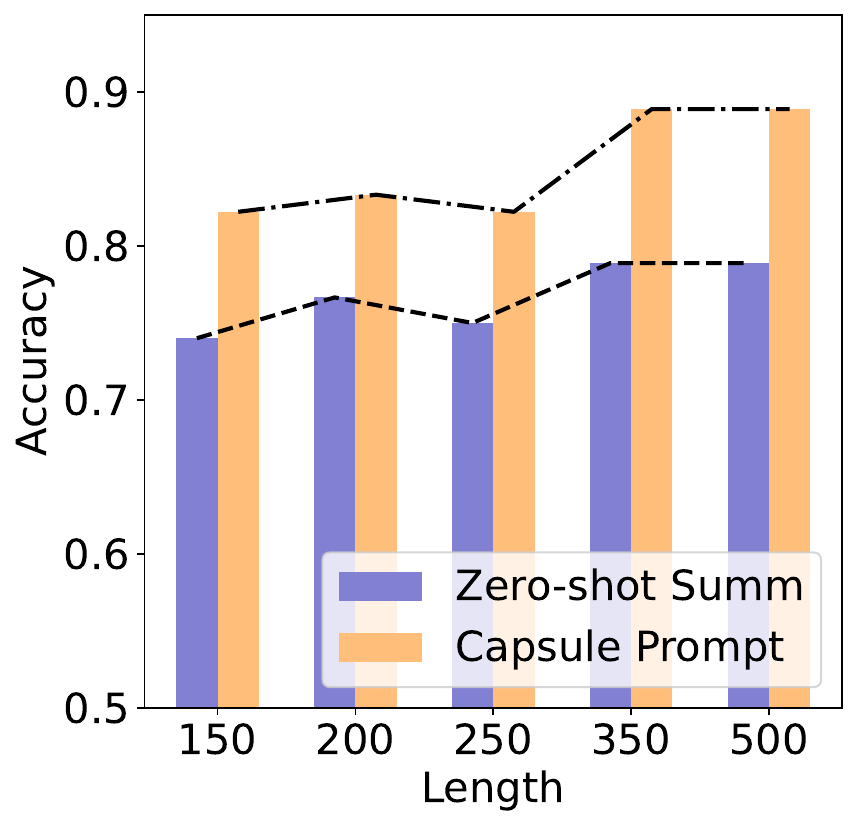}
\endminipage\hfill
\minipage{0.5\columnwidth}
  \includegraphics[width=1.0\columnwidth]{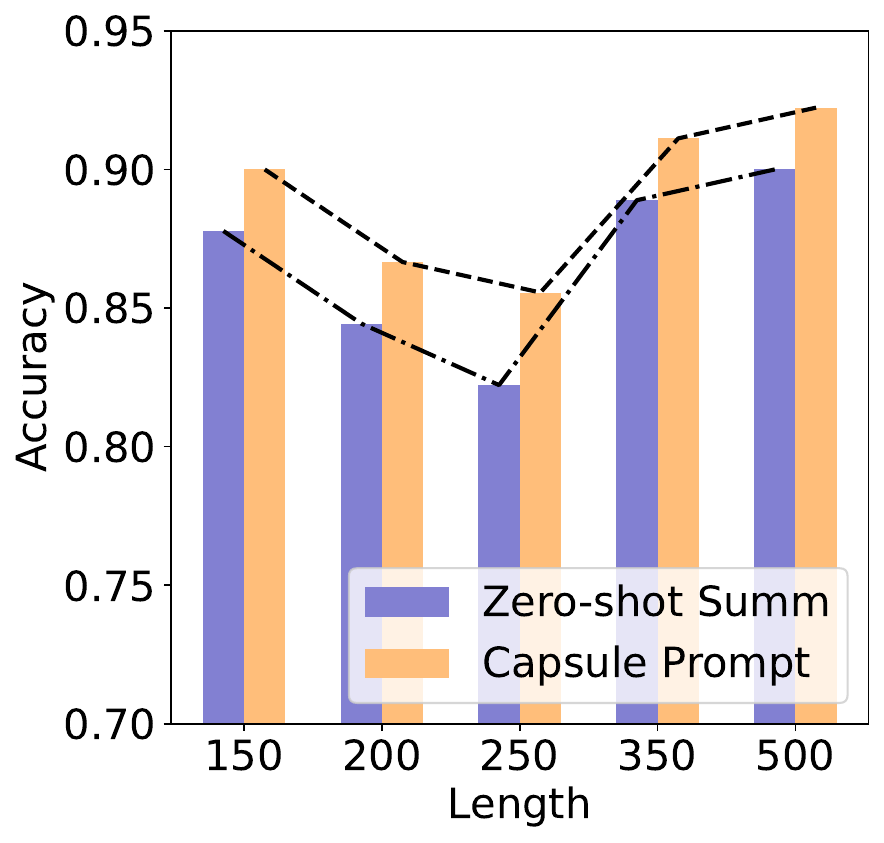}
\endminipage
\vspace{-0.2cm}
\caption{Impact of prompt length on Vicuna-13B (left) and Claude2 (right) on TriviaQA dataset.}
\vspace{0.15cm}
\label{fig:length}
\end{figure}

\textit{Secondly}, we carry out ablation studies to establish the effectiveness of the reward function in \Algnameabbr{}. These studies are conducted on Vicuna-13B and Claude2 using the TriviaQA-Long dataset. As shown in Figure~\ref{fig:gptsumm} (right), the results indicate a degradation in performance for downstream tasks when the reward function is not utilized. Note that "w/ reward" means Nano-Capsulator trained with the reward function, while "w/o reward" denotes a Nano-Capsulator trained without the reward function. 
We further present case studies of \Algname{} on GSM8K to showcase the logic preservation, as illustrated in Figure~\ref{fig:gsmcase} from Appendix~\ref{apx:demo}. These studies clearly demonstrate that \Algname{} retains more semantic meanings by preserving complete logical structures, suggesting that the utility of prompts is better maintained.

\begin{figure}[t]
\minipage{0.5\columnwidth}
  \includegraphics[width=1.0\columnwidth]{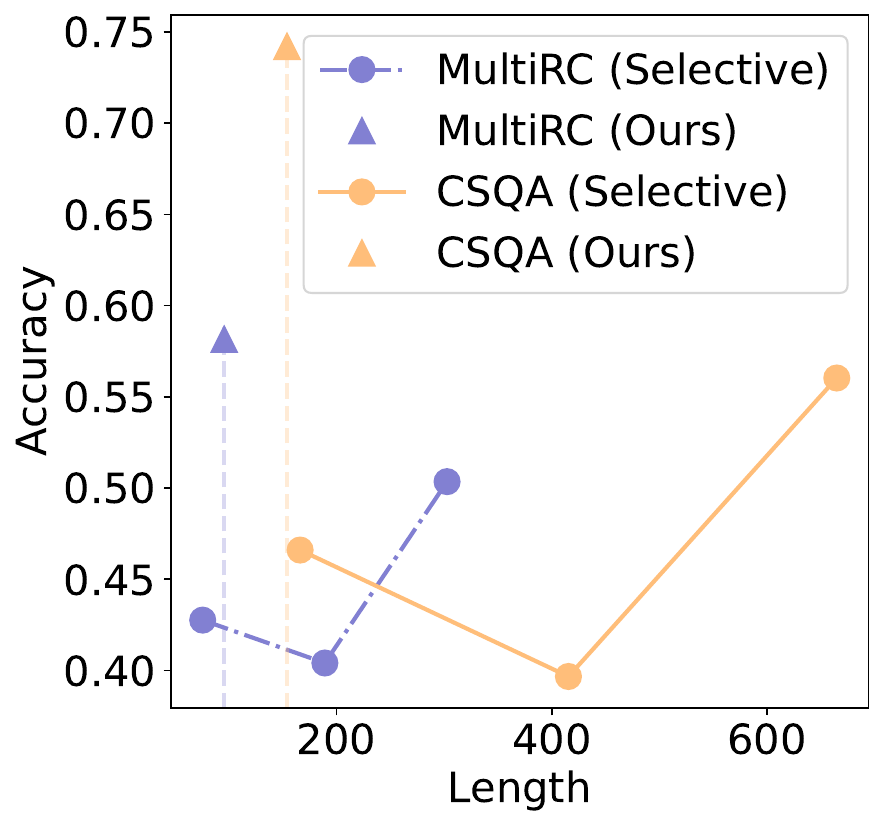}
\endminipage\hfill
\minipage{0.5\columnwidth}
  \includegraphics[width=1.0\columnwidth]{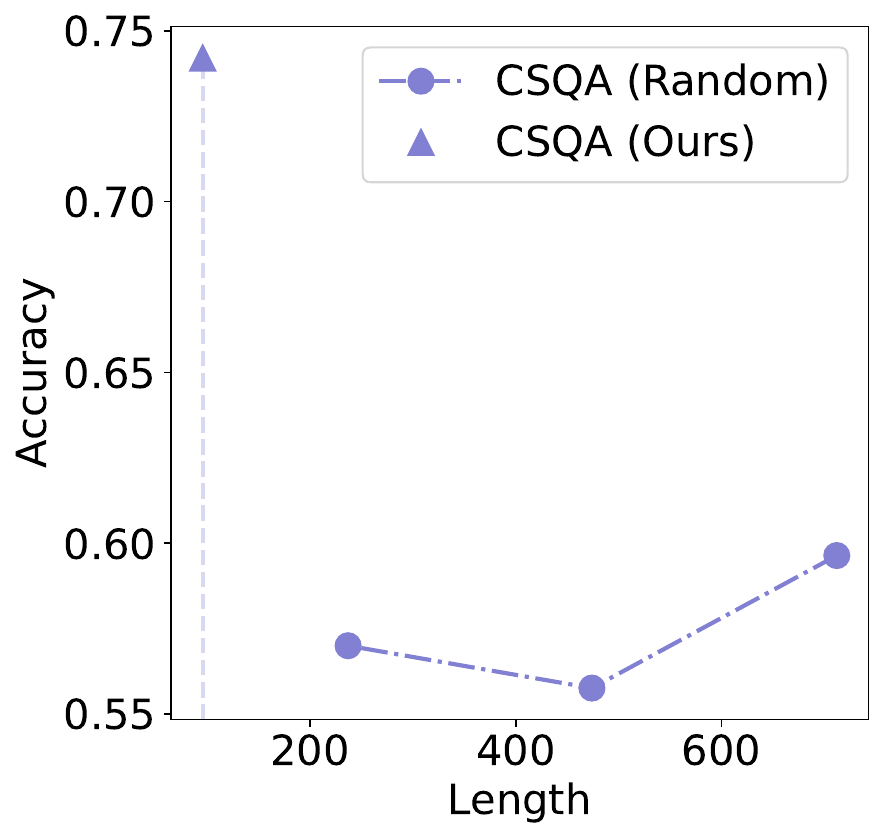}
\endminipage
\caption{The comparison results of \Algname{} and text dropping methods, including Selective Context (left) and random demonstration elimination (right).}
\vspace{-0.1cm}
\label{fig:del}
\end{figure}

\subsection{Exploration of Impact Factors (RQ3)}
In this section, we explore our proposed compression mechanism in greater detail, examining various factors that influence its performance.

\begin{figure}[t]
\minipage{0.5\columnwidth}
  \includegraphics[width=0.98\columnwidth]{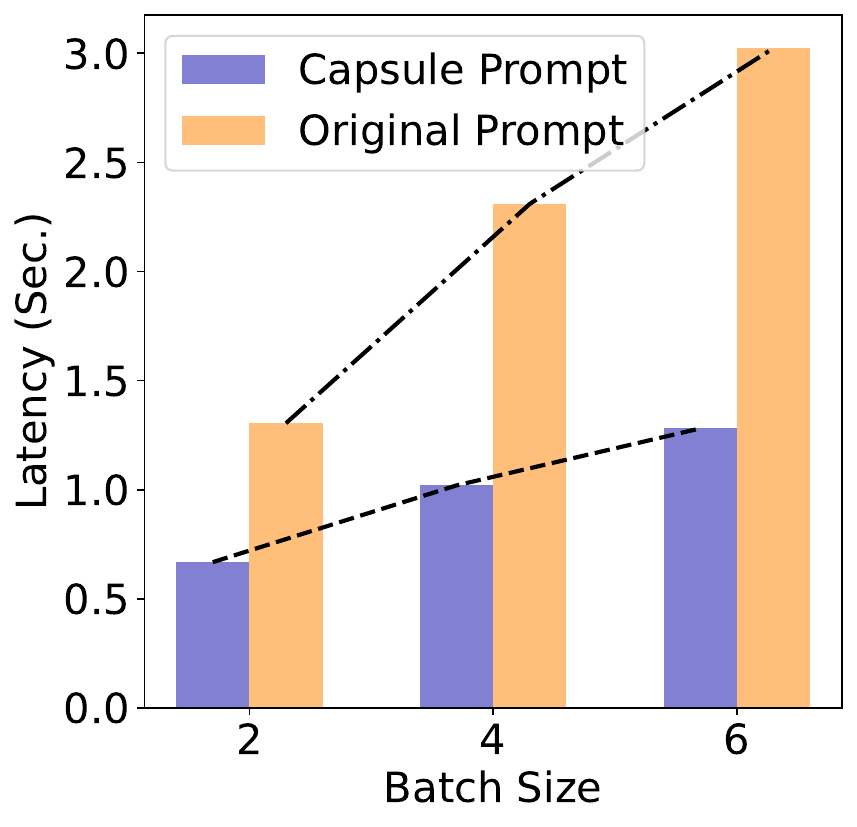}
\endminipage\hfill
\minipage{0.5\columnwidth}
  \includegraphics[width=0.95\columnwidth]{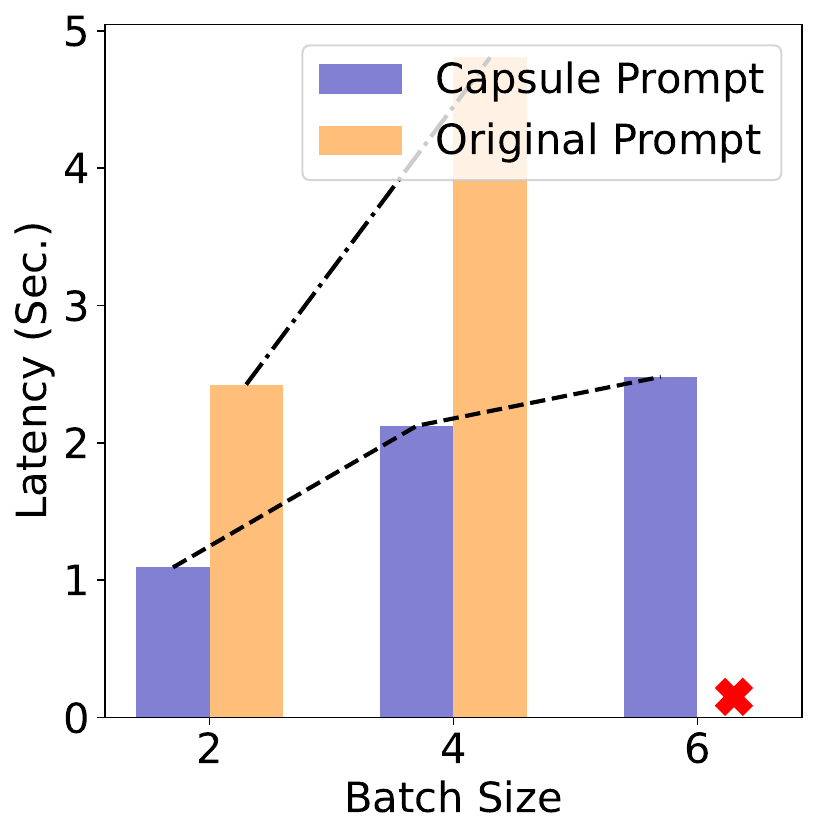}
\endminipage
\caption{Inference Latency of Vicuna-13B on CSQA dataset (left) and TrivaQA-Long dataset (right), where \cmark~ indicates out of memory.}
\vspace{-0.2cm}
\label{fig:vcu_speed}
\end{figure}

\vspace{0.2cm}
\noindent\textbf{Impact of Capsule Prompt Length.} In the main experiment, the length constraint is fixed to 150 or 300 tokens for the prompt compression.
We further explore how the compression rate affects the preservation of utility. The results are displayed in Figure~\ref{fig:length}, obtained from experiments conducted using the TriviaQA dataset with Vicuna-13B and Claude2. We observe that the length of \Algname{} can impact its utility on different LLMs. While longer prompts might capture more useful information for downstream tasks, they can also introduce certain noise or misinformation to the LLMs. This can result in suboptimal performance when specific LLMs interact with the compressed prompts. The situation can be observed from the results of \Algname{} and Zero-shot Summ Prompt, where the length of 150 outperforms other length settings under Claude2 and the length of 200 outperforms other length settings Vicuna-13B. We notice that the desired length settings of \Algnameabbr{} can be observed from the performance of the Zero-shot Summ Prompt, as they share similar performance trends. 

\vspace{0.2cm}
\noindent\textbf{Impact of Discrete Text Elimination.}
In addition to compressing prompts using \Algnameabbr{}, we acknowledge that prompt length can also be reduced by employing methods like random dropping or rule-based selection.
To this end, we have conducted studies comparing the performance of straightforward text dropping with our proposed framework. We consider two baseline methods under Claude2: a naive random demonstration elimination on the CSQA dataset and Selective Context as described in~\cite{li2023compressing}, in which Selective Context eliminated the word according to the self-information values, on both the CSQA and MultiRC datasets. 
The outcomes of these comparisons are showcased in Figure~\ref{fig:del}. We observe that \Algname{} outperforms the other two baselines. Particularly, \Algname{} achieves better performance when the length is similar to the baselines. This again demonstrates the effectiveness of \Algname{} in preserving the utility.

\subsection{Latency of \Algnameabbr{} (RQ3)} 
\begin{figure}[t]
\minipage{0.5\columnwidth}
  \includegraphics[width=1.0\columnwidth]{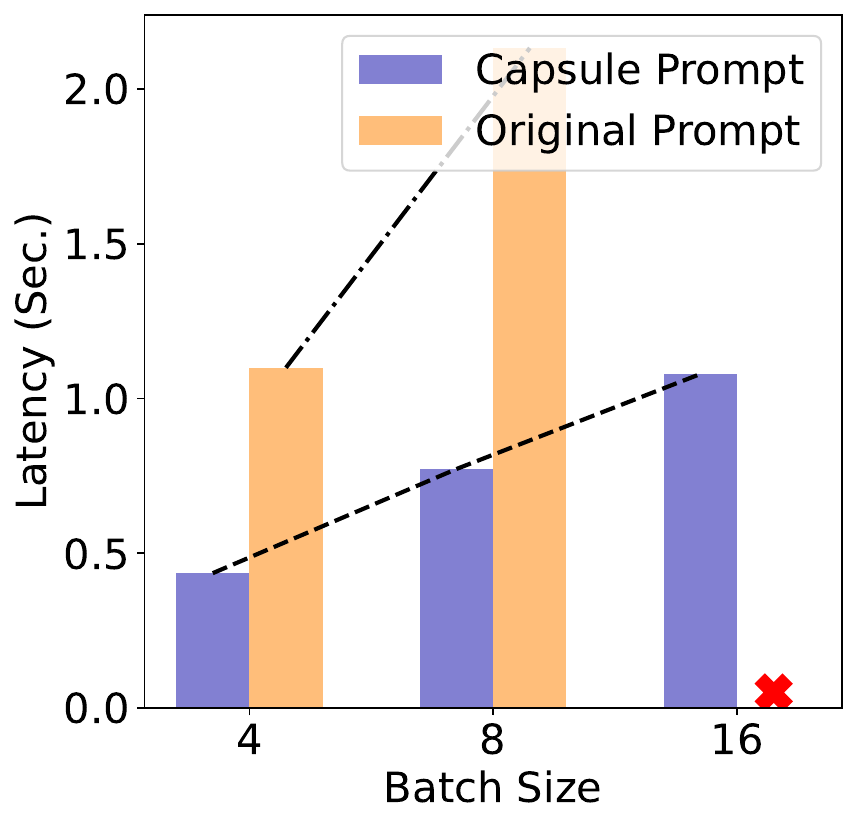}
\endminipage\hfill
\minipage{0.5\columnwidth}
  \includegraphics[width=1.0\columnwidth]{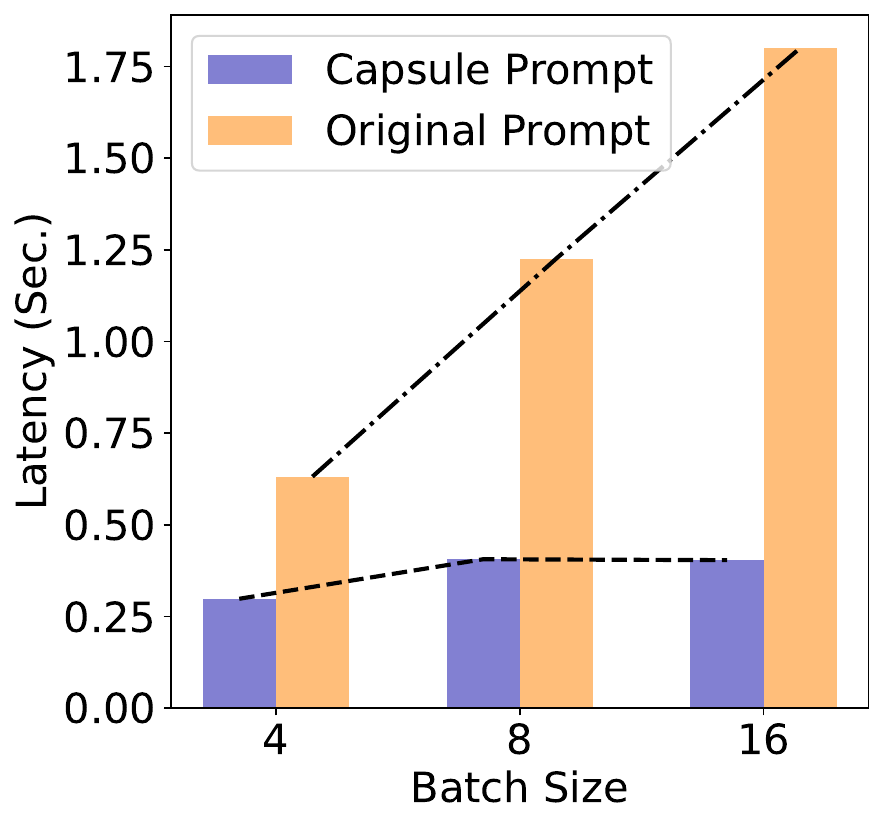}
\endminipage
\caption{Inference Latency of OPT-2.7B on CSQA dataset (left) and TrivaQA-Long dataset (right), where \cmark~ indicates out of memory.}
\vspace{-0.2cm}
\label{fig:opt_speed}
\end{figure}

The configuration of the computational infrastructure is given in Appendix~\ref{apx:infra}. We conducted the efficiency experiments on two publicly available LLMs: OPT-2.7B~\cite{zhang2022opt} and Vicuna-13B~\cite{vicuna2023}, under different batch size settings with the generated length of $200$ tokens. Due to the limited GPU memory, we set Vicuna-13B as bfloat16 to accommodate a single GPU, while OPT-2.7B remains its inherent version. As demonstrated in Figure~\ref{fig:vcu_speed} and Figure~\ref{fig:opt_speed},
we observe that \Algnameabbr{} framework achieves much lower computational latency compared to the case while inputting the original prompt on both LLMs. Specifically, \Algname{} obtains mostly of its original performance while reducing $2.1\times \sim 4.5\times$ of execution latency. As depicted in Figure~\ref{fig:opt_speed}, \Algname{} is capable of being accommodated within OPT-2.7B under a larger batch size, whereas the use of the original longer prompt leads to an out-of-memory issue (i.e., when batch size = 16). Additionally, we notice that \Algname{} achieves considerable benefits in speeding up the inference process as the batch size increases. Under OPT-2.7B, \Algname{} accelerates the process by up to $4.5\times$, and under Vicuna-13B, it achieves a speed increase of $4.1\times$ compared to the original input prompt. This indicates that \Algname{} allows for a larger batch size while reducing the time required for the inference process.

\section{Conclusion}
Our work introduces \Algnameline{} (\Algnameabbr{}), a framework for effectively compressing long prompts for LMs while preserving essential information. \Algnameabbr{} alleviates the context length limitations of LLMs, enhancing processing efficiency and cost-effectiveness. Our results show that \Algnameabbr{} reduces prompt lengths by 81.4\%, decreases inference latency by up to 4.5$\times$, and cuts budget overheads by 80.1\% with almost identical accuracy and relevance. 
This demonstrates its significant potential for improving LLM efficiency across various applications that utilize long input documents. Future research will focus on refining \Algnameabbr{} for broader domain applications and exploring its usage in data-intensive fields.

\newpage
\bibliography{paper}
\bibliographystyle{acl_natbib}

\appendix

\vspace{0.3cm}
\section*{Appendix}
\section{Computation Infrastructure}
\label{apx:infra}
For a fair comparison of testing algorithmic throughput, the experiments are conducted based on the following physical computing infrastructure in Table~\ref{tab:computing_infrastructure}. 

\begin{table}[h!]
\centering
\begin{tabular}{l|c}
\hline
Device Attribute & Spec \\
\hline\hline
Computing infrastructure & GPU \\
GPU model & Nvidia-A40 \\
GPU number & 1 \\
GPU Memory & 46068 MB \\
\hline
\end{tabular}
\caption{Computing infrastructure for the experiments.}
\label{tab:computing_infrastructure}
\vspace{-3mm}
\end{table}

\section{Additional Experiments of Comparison to Soft Prompt Baselines}
\label{apx:demo}
To evaluate our proposed framework against existing soft prompt methods, we conduct experiments with AutoCompressors~\cite{chevalier2023adapting} on the GSM8K dataset, as shown in Table~\ref{tab:adpt}. Our Capsule Prompt is utilized for predictions using the Llama-2-7B model, which is identical to the pre-trained model used by AutoCompressor. As we can see, AutoCompressor does not preserve essential information in the compressed soft prompts, leading to a considerable performance drop in the GSM8K task.

\begin{table}[h!]
\centering
\begin{tabular}{l|c|c}
\hline
GSM8K & AutoCompressors & Ours \\
\hline\hline
Accuracy & 3.79 & 19.7 \\
\hline
\end{tabular}
\caption{Computing infrastructure for the experiments.}
\label{tab:adpt}
\vspace{-3mm}
\end{table}

\section{API Cost of PaLM}
We here provide the API cost of PaLM during the evaluation of \Algnameabbr{}. We can observe that \Algname{} generated by \Algnameabbr{} save up to 80.1\% of its original cost on PaLM. The results further underscore the excellent cost-efficiency attributes of \Algname{}.

\label{apx:cost}
\begin{table}[t]
\small
\centering
	\begin{tabular}{l ccc}
        \toprule
        & \multicolumn{3}{c}{PaLM~\cite{vertex}} \\
        \cmidrule(lr){2-4}
        Cost(\$) & Original & \Algname{} & Saved \\
        \midrule 
        CSQA & 0.156 & 0.034 & -77.9\% \\
        GSM8K & 0.054 & 0.019 & -63.9\% \\
        MultiRC & 0.478 & 0.135 & -71.6\% \\
        TrivaQA-Long & 0.022 & 0.004 & -80.1\% \\
        \bottomrule
    \end{tabular}%
    \caption{API cost comparison of \Algname{} and original prompt on PaLM, where \Algname{} save up to 80.1\% of the original cost.}
\label{tab:palm_cost}
\vspace{-0.3cm}
\end{table}

\section{Training Costs of \Algnameabbr{}}
In this section, we discuss the training time the memory cost of \Algnameabbr{} in this section. All datasets are trained using the initial weights of Vicuna-7B. Training time and memory requirements are different with the volume and types of training data. In our work, the training time for the few-shot CoT compression task is approximately 8 hours, and for the reading comprehension compression task, it is about 4 hours. Once the \Algnameabbr{} completes the training process, we can directly derive the compressed hard prompts through a single forward pass of \Algnameabbr{}.

\section{The Case Studies of \Algname{}}
\label{apx:demo}
\begin{figure*}[t!]
    \centering
    \includegraphics[width=.95\textwidth]{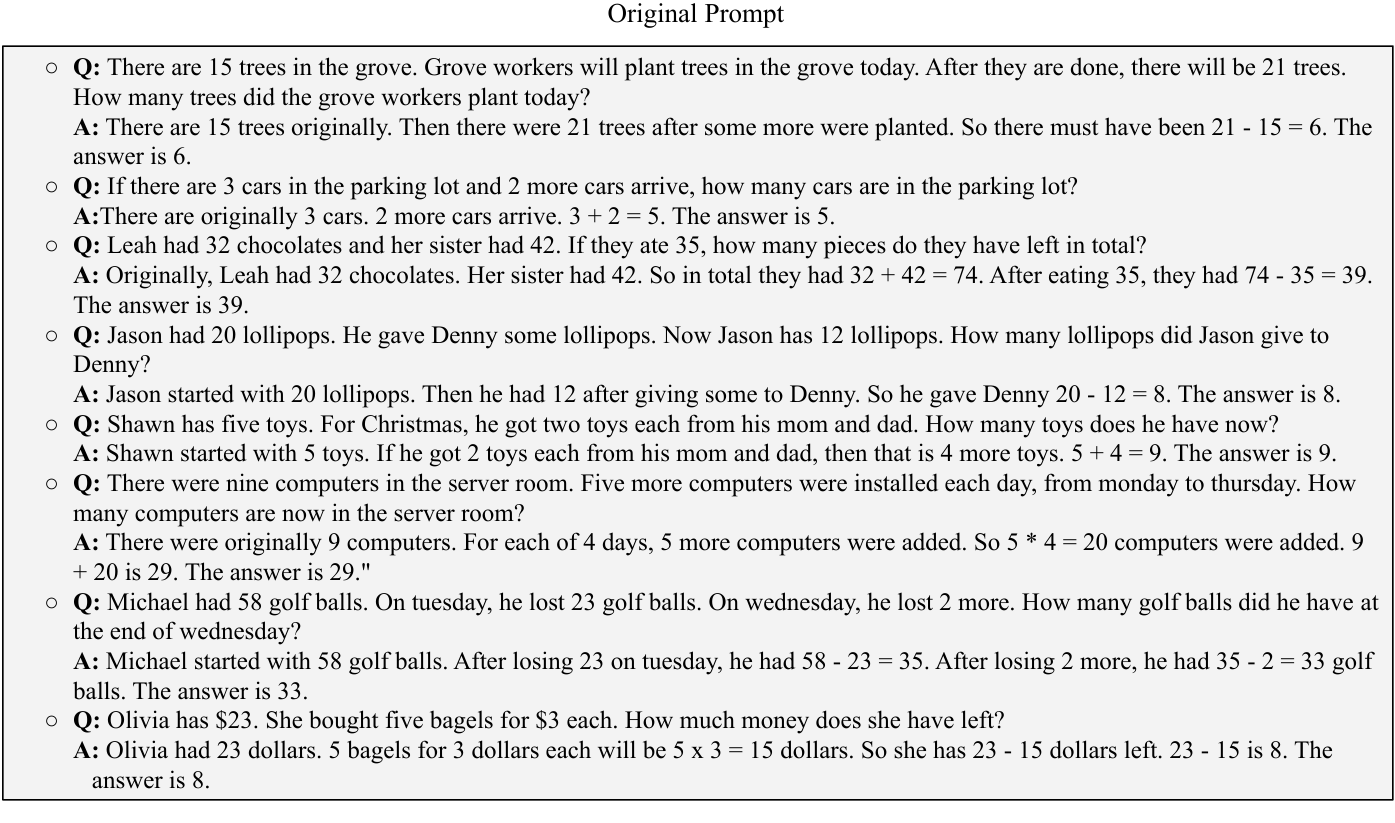}
    \vspace{0.5cm}
    \includegraphics[width=.95\textwidth]{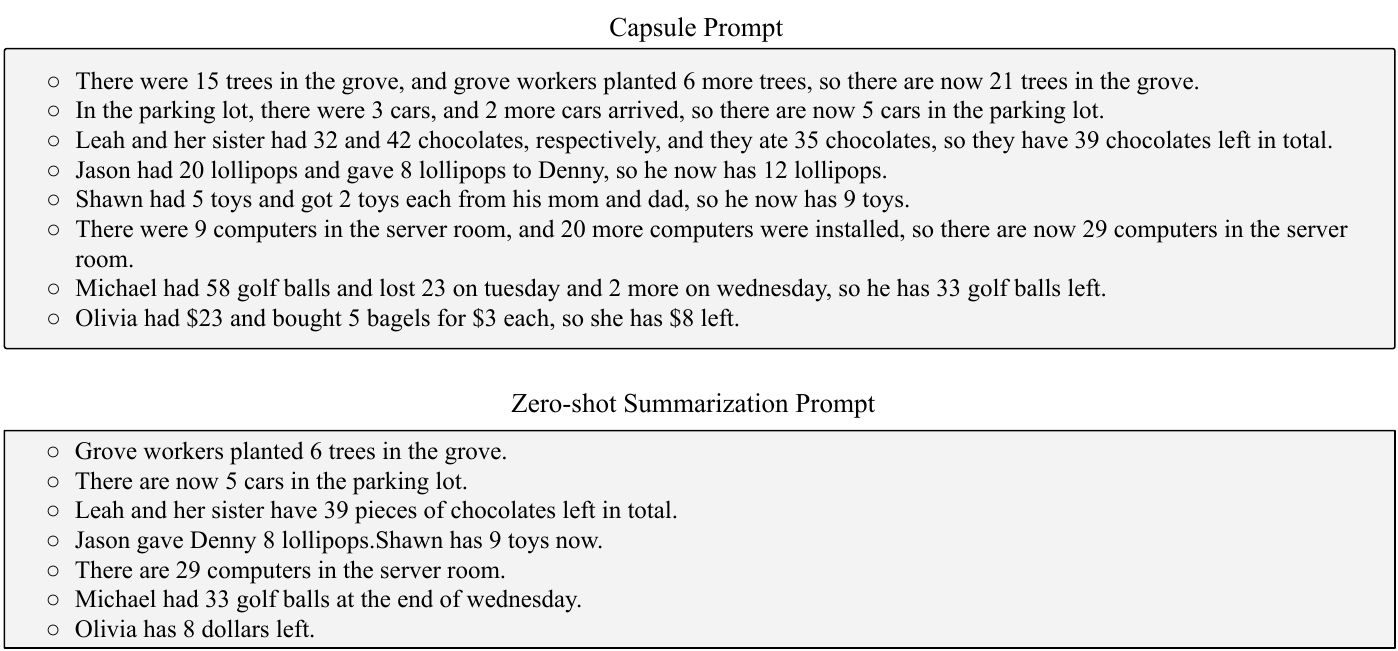}
    \caption{A case study on GSM8K dataset. The results are the \Algname{} and in-context summarization prompt generated Vicuna-7B, following the settings in RQ2.}
    \label{fig:gsmcase}
\end{figure*}

\begin{figure*}[t!]
    \centering
    \includegraphics[width=.95\textwidth]{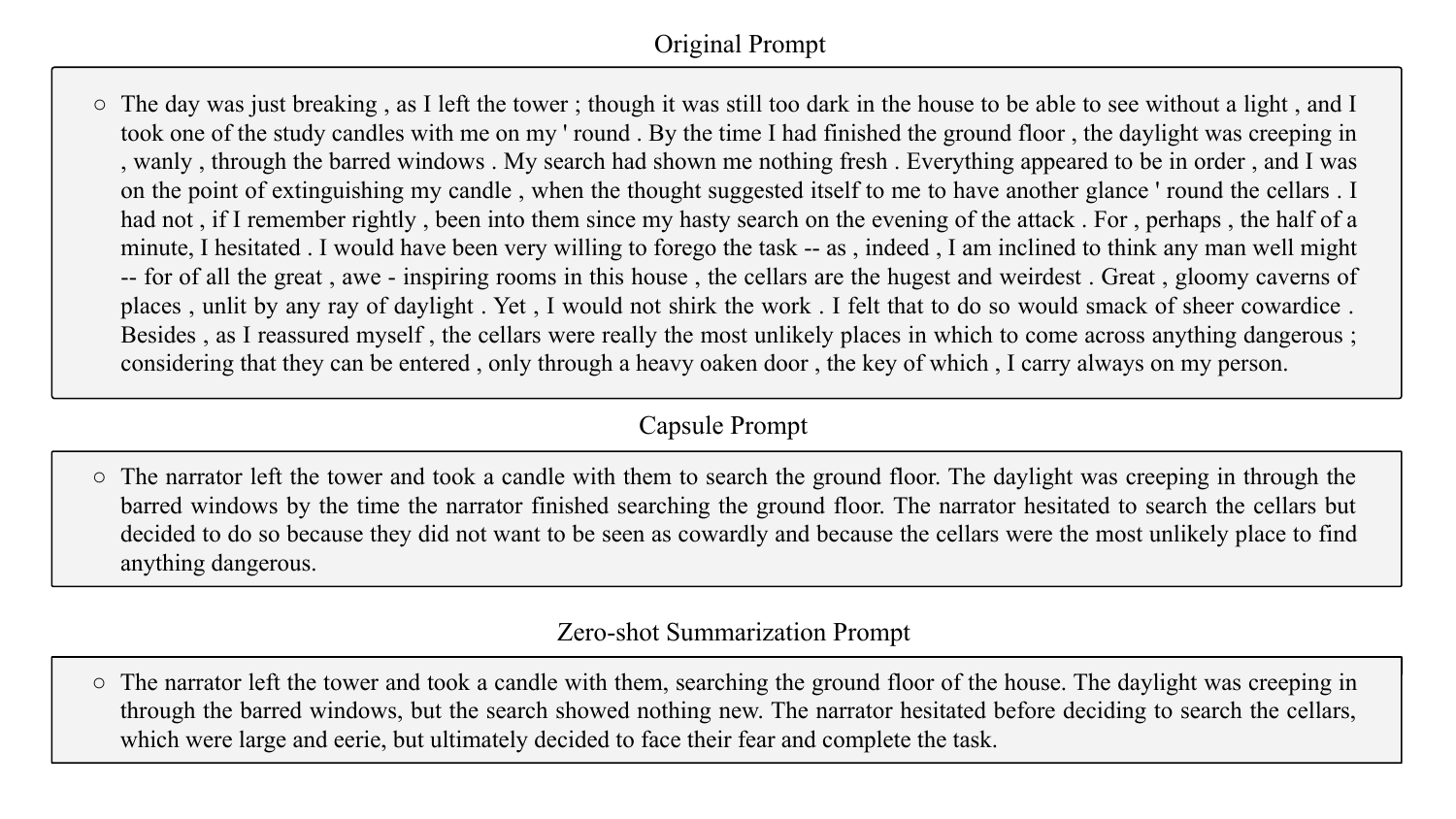}
    \caption{A case study on MultiRC dataset. The results are the \Algname{} and in-context summarization prompt generated Vicuna-7B, following the settings in RQ2.}
    \label{fig:mltrcase}
\end{figure*}

We demonstrate the case study of \Algname{} on the GSM8K and MultiRC dataset. The results, depicted in Figure~\ref{fig:gsmcase} and Figure~\ref{fig:mltrcase}, show that \Algnameabbr{} can better obtain the semantic meanings of math logic from the original CoT prompt. We observe that in the absence of \Algnameabbr{}, the Zero-shot Summarization approach struggles to compress long prompts effectively.

\section{Instruction Usage in Inference LLMs}
We provide a list of the instruction utilized in training our \Algnameabbr{} framework in Table~\ref{tab:instruct}, including $\textbf{T}_{\text{Rep}}$ replicating instruction and summarizing instruction $\textbf{T}_{\text{Summ}}$.
\label{apx:prompt}

\begin{table*}[t!]
    \small
    \centering
    \begin{tabular}{l|p{4.5cm}|p{7cm}} \toprule
        Compression Tasks & $\textbf{T}_{\text{Rep}}$ (Replicating Instruction) & $\textbf{T}_{\text{Summ}}$ (Summarizing Instruction) \\ 
        \toprule\toprule
        Few-shot CoT & Repeat the following main input. & Please summarize each question-answer pair in one sentence within less than \{word count\} words. Make sure not to repeat the input question-answer pair. \\\toprule
        Reading Comprehension & Repeat the following main input. & Please summarize the passage within less than \{word count\} words. Make sure not to repeat the passage. \\
    \bottomrule
    \end{tabular}
    \caption{Instructions used in training \Algnameabbr{}.}
    \label{tab:instruct}
\end{table*}

\end{document}